\documentclass{ws-aa_mod}

\usepackage{amssymb}
\usepackage{graphicx}
\usepackage{xcolor}
\usepackage{comment}

\newcommand{\R}{\mathbb{R}}

\newcommand{\N}{\mathbb{N}}

\newcommand{\A}{\boldsymbol{\alpha}}

\newcommand{\act}{\phi}
\newcommand{\dist}{\mathrm{Dist}}
\newcommand{\rank}{\mathrm{Rank}}

\newcommand{\conv}{\mathrm{CH}}

\begin{document}

\markboth{H.-P. Beise and S. Dias Da Cruz}{Basin of attraction of autoencoders}

%
\catchline{}{}{}{}{}
%

\title{Topological properties of basins of attraction of width bounded autoencoders}

\author{Hans-Peter Beise}

\address{ Department of Computer Science\\
    Trier University of Applied Sciences\\
Trier, 54293,
Germany\\
beise@hochschule-trier.de}

\author{Steve Dias Da Cruz}

\address{Basics and Mathematical Models, IEE S.A.\\
1, rue du Campus, Bissen, 7795, Luxembourg\\
steve.dias-da-cruz@iee.lu}

\maketitle

\begin{history}
\received{(Day Month Year)}
\revised{(Day Month Year)}
\end{history}

\begin{abstract}
  In \cite{radhakrishnan2020overparameterized}, the authors empirically show that autoencoders trained with standard SGD methods form a basins of attraction around their training data. We consider network functions of width not exceeding the input dimension and prove that in this situation, such basins of attraction are bounded and their complement cannot have bounded components. Our conditions in these results are met in several experiments reported in \cite{radhakrishnan2020overparameterized} and we thus address a question posed therein. We also show that under some more restrictive conditions, the basins of attraction are path-connected. The necessity of the conditions in our results is demonstrated by means of examples. 
\end{abstract}



\section{Introduction}

In the context of artificial neural networks, the term \textit{associative memory} describes the ability of certain networks to allow the retrieval of ``memorized" data via the activation of associated features. In mathematical terms this means that a certain iterative procedure converges to some limit which is interpreted as memorized pattern.
The research on associative memory in this context has a long history dating back to the early seventies \cite{anderson1972simple,kohonen1972correlation,nakano1972associatron} and is mainly associated with network types known as Hopfield networks \cite{hopfield1982neural}. The theoretical foundations of Hopfield networks have been intensively studied and, as for instance, their capacity to store data and convergence of data retrieval are understood to a large extent \cite{mceliece1987capacity,bruck1990convergence}. For recent developments we refer to \cite{krotov2016dense,demircigil2017model, ramsauer2020hopfield}.
An autoencoder, on the other hand, is a type of neural network that, in its original form, is designed to learn an efficient data encoding. For this purpose it generates an encoding of the data and reconstructs them from the latter simultaneously during training. By now, there exists a huge number of different types of autoencoders and they have been omnipresent in deep learning applications in recent years. We refer to \cite{bank2020autoencoders,goodfellow2016deep} and the references therein.
In \cite{radhakrishnan2020overparameterized}, the authors report the empirical finding that autoencoders trained  with standard stochastic gradient descent (SGD) methods until the training loss vanishes to zero effectively function as an associative memory of their training data. In a series of experiments they demonstrate that the iterative application of a trained autoencoder on (perturbed) input data very robustly converges to an example from the training data. That is, for an autoencoder $F$ and (many) training points $x^*$, it is observed that 
$F^k(x)\rightarrow x^*$ as $ k\rightarrow \infty,$
where $F^k$ means the $k$-fold iterative application of $F$, for all $x$ in some subset $B(x^*)$ of the input space, called \textit{basin of attraction}. These training examples hence constitute attractive points of the autoencoder and can be retrieved by iterative application of $F$. These findings reveal a bias away from an approximation of the identity in the vicinity of the training data towards basins of attraction. A similar observation was made in \cite{zhang2019identity}, in which the authors show that different architectures exhibit different biases between the extreme cases of the identity function and constant functions when trained to reconstruct the training data. Other works addressing the inductive biases of gradient descent in deep learning include \cite{neyshabur2014search,soudry2018implicit,gunasekar2018implicit, belkin2019reconciling}. 
The effect of attractive fixed points learned by autoencoders, as observed in \cite{radhakrishnan2020overparameterized}, is also further investigated in \cite{jiang2020associative}. In the latter work, the authors derive theoretical results on a mechanism that can explain the occurrence of attractive fixed points of certain autoencoders in the infinite width limit.

The presence of multiple isolated fixed points, as observed in \cite{radhakrishnan2020overparameterized}, is made possible by the utilization of a non-linear activation function. Without such non-linearities, common neural network functions, as considered in \cite{radhakrishnan2020overparameterized} and in this work, would be reduced to affine mappings.
In our review of the available literature, we have not come across any theoretical findings pertaining to the existence of basins of attraction of autoencoders.

In this work we consider autoencoders having widths that do not exceed the input dimension, and study topological properties of their basins of attraction. 
Our research is related to investigations on topological properties of subsets of the input space or approximation properties of neural networks in a bounded width setting \cite{johnson2018deep,hanin2017approximating,beise2018decision,grigsby2022transversality,nguyen2018neural}.

The contributions of this work are summarized next:
\begin{itemize}
    \item In Theorem \ref{attractor_unbounded}, we show that for autoencoders in the bounded-width setting mentioned above, equipped with a continuous, monotonically increasing activation function, and such that at least one weight matrix has a rank strictly smaller than the dimension of the input data space, each component of a basin of attraction is unbounded. 
    \item In Theorem \ref{attractor_doesnot_enclosed_points}, it is demonstrated that under the before-mentioned assumption, except for the condition on the rank of the weight matrices, the complement of a basin of attraction of such an autoencoder cannot have bounded components.
    
    \item Example \ref{ex:RankCond} and Example \ref{ex:unboundBasin} show that the assumptions in Theorem \ref{attractor_unbounded} and Theorem \ref{attractor_doesnot_enclosed_points} are necessary or cannot be dropped without substitute.

    \item In Example \ref{UnconBasinEx}, we construct a six-layer neural network function that satisfies the aforementioned width condition but has a non-path-connected basin of attraction. This construction becomes possible by employing a non-surjective activation function, as is subsequently established in Theorem \ref{attractor_pathconnected}. This theorem states that for autoencoders with square, full-rank weight matrices and that are endowed with a continuous, monotonically increasing, and surjective activation function, such as leaky ReLU, every basin of attraction is path-connected.

\end{itemize}

\subsection{Notation}\label{sec:notations}
Let us introduce the following notation: Let $\left\Vert \cdot\right\Vert$ denote the Euclidean norm, and for $x\in\R^n$, let  $U_\varepsilon(x):=\{y\in\R^n:\Vert x-y\Vert<\varepsilon\}$, where $\varepsilon>0$. For a set $D\subset \R^n$, we denote by $D^\circ$ the set of interior points. By $\overline{D}$ we mean the closure of $D$ and $\partial D=\overline{D}\setminus D^\circ$ denotes its boundary. The application of a mapping $f$ on set $M$ means $f(M):=\{f(x):x\in M\}$. The $k$-fold application of a mapping $f:\R^n\rightarrow \R^n$ is denoted by $f^k$, i.e. $f^k=f\circ f^{k-1}$ for $k>1$ and $f^1=f$. For $x_1,x_2\in \R^n$, we denote by $\conv(x_1,x_2)$ the convex hull of these points, which is equal to the line segment that connects $x_1$ and $x_2$. For $x\in\R^n$ and $M\subset \R^n$ let $\dist(x,M):=\inf_{y\in M}\Vert x-y\Vert$.
For $v\in\R^n$ and $j\in\{1,...,n\}$, we write $v_{(j)}$ for the $j$th coordinate of $v$. The identity matrix is denoted by $I_n\in\R^{n\times n}$ and $\rank(A)$ denotes the rank of a matrix $A\in\R^{m\times n}$. For functions of one scalar variable $f:\R\rightarrow \R$, the application on some $x\in\R^n$ is defined coordinate-wise: $f(x)=(f(x_{(1)}),...,f(x_{(n)}))^T$. The inverse image $f^{-1}(D)$ for $D\subset \R^n$ is defined accordingly. 
To avoid confusions in formulations of the results and the proofs, let us also clarify that a real-valued function is said to be monotonically increasing when $x<y$ implies $f(x)\leq f(y)$, and it is said to be strictly monotonically increasing when $x<y$ implies $f(x) < f(y)$.
We also recall that $C\subset \R^n$ is called \textbf{path-connected} if for every pair of points $x_1,x_2\in C$, there exists a continuous path $\gamma:[0,1]\rightarrow C$ such that $\gamma(0)=x_1$ and $\gamma(1)=x_2$. A set $C$ is said to be \textbf{connected} if there are no disjoint open sets $U,\, D  \subset \R^d$ such that $C\subset U\cup D$, and $C\cap U\neq \emptyset$ and $C\cap D \neq \emptyset$. For a given $M\subset \R^n$, $C\subset M$ is said to be a (connected) component of $M$, when $C$ is connected and every $\tilde{C}\subset M$ with $C\subset \tilde{C}$ and $\tilde{C}\setminus C\neq \emptyset$ is not connected.

We consider neural network functions $F:=F_L:\R^{n_0}\rightarrow \R^{n_L}$, recursively defined by
\begin{equation}\label{networkFun}
	F_k(x):=\begin{cases} 
	\act(W_k x+b_k)&\mathrm{for}\  k=1\\
	\act(W_k F_{k-1}(x)+b_k)\ &\mathrm{for}\  k\in\{2,...,L-1\}\\
    W_k F_{k-1}(x)+b_{k} &\mathrm{for}\  k=L.
	\end{cases}
\end{equation} 
In the above definition, we have $W_j\in \R^{n_{j}\times n_{j-1}}$ (weights), $b_j\in \R^{n_j}$ (bias), for $j=1,...,L$, and $\act:\R \rightarrow \R$, called the activation function. Furthermore, let
\begin{equation}\label{networkFunLin}
    \A_j(x):=W_jx+b_j, \ j=1,...,L.
\end{equation}
We call $n_j$ the \emph{width} of layer $j=1,...,L$. The width of the network is defined as $\max\{n_j: \, j=1,...,L\}$ and $L$ is called the depth of the network.

Let us further recall the following definition from dynamical systems, c.f. \cite{strogatz2018nonlinear}. 

\begin{definition}\label{defAttPoint}
A point $x^*\in\R^n$ is called an \textbf{attractive point} if there exists an open neighbourhood $\mathcal{O}$ of $x^*$ such that for all $x \in \mathcal{O}$, $f^{k}(x)\rightarrow x^*$ as $k\rightarrow \infty$.
\end{definition}
As the functions considered in the sequel are continuous, we will have that every attractive point $x^*$ must be a fixed point, i.e. $f(x^*)=x^*$. We focus solely on isolated fixed points, as this is observed to be the prevailing case in (non-linear) autoencoders \cite{radhakrishnan2020overparameterized}. 
\begin{definition}
Let $x^*$ be an attractive point, then the set of points $x$ such that  $f^k(x)\rightarrow x^*$ as $k\rightarrow \infty$ is called the $\textbf{basin of attraction}$ of $x^*$ and is denoted by $B(x^*)$. 
\end{definition}

\section{Topological properties of basins of attraction}\label{topPropertiesSec}

In this section we present our main results. We consider neural network functions $F:\R^{n_0}\rightarrow \R^{n_0}$, called \textbf{autoencoders}. In applications, autoencoders are usually trained to approximate the identity in the sense that $\Vert F(x)-x\Vert$ is small (or the deviation in terms of some other distance) on a certain subset in $\R^{n_0}$. 
As already mentioned in the introduction, it has been discovered in \cite{radhakrishnan2020overparameterized} that training with standard SGD-type methods until the training loss vanishes to zero leads to autoencoders that form basins of attraction with training examples as attractive points in the input space.

\subsection{Unboundness of basins of attraction}

\begin{theorem}\label{attractor_unbounded}
Let $F$ be an autoencoder of width at most $n_0$  with continuous, monotonically increasing activation function and assume that \begin{equation}\label{rankCond}
   \min \{\rank(W_j):j=1,...,L\}<n_0.  
\end{equation}If $x^*$ is an attractive point, then every component of $ B(x^*)$ is unbounded.
\end{theorem}
Note that $B(x^*)$ in the last result can be disconnected, c.f. Example \ref{UnconBasinEx} and the subsequent comments.
\begin{theorem}\label{attractor_doesnot_enclosed_points}
Let $F$ be an autoencoder of width at most $n_0$ and with continuous, monotonically increasing activation function. If $x^*$ is an attractive point, then $\R^{n_0}\setminus B(x^*)$ has no bounded component.
\end{theorem}

The results presented in this subsection share similarities with results found in the context of approximation properties of networks functions having width at most $n_0$. For the case of one-to-one activation functions, it is proven in \cite{johnson2018deep} that level sets for real-valued neural networks having width at most $n_0$ are unbounded, c.f. \cite[Lemma 4]{johnson2018deep}. A similar result for the case of ReLU activation is given in \cite{hanin2017approximating} in the context of their proof of the lower bound of \cite[Theorem 1]{hanin2017approximating}.
The following lemma constitutes an important tool in the proofs of Theorem \ref{attractor_unbounded} and Theorem \ref{attractor_doesnot_enclosed_points}. In essence, it enables us to take our conclusions for the case of monotonic activation functions instead of strictly monotonic activation functions.
If we would restrict Theorem \ref{attractor_unbounded} or Theorem \ref{attractor_doesnot_enclosed_points} to strictly monotonically increasing activation functions, arguments as they are used in the proof of \cite[Lemma 4]{johnson2018deep} could replace the use of Lemma \ref{lemma_nottoouter} in the respective proofs of Theorem \ref{attractor_unbounded} or Theorem \ref{attractor_doesnot_enclosed_points}. 

\begin{lemma}\label{lemma_nottoouter}
Let $\act:\R\rightarrow \R$ be a continuous and monotonically increasing function, and $D\subset \R^n$ a bounded set. Then $\partial \act(D)\subset \act(\partial D)$.
\end{lemma}
Note that $\partial \act(D)= \act(\partial D)$ in the previous result does not hold true in general. Also note that the assertion of the lemma is trivial when $\act$ is strictly monotonically increasing.

\begin{proof}
We show that $y\in \partial \act(D)$ implies $y\in\act(\partial D)$. To this end let $(y_m)$ be a sequence in $\act(D)$ with $y_m\rightarrow y$ as $m\rightarrow\infty$.
For every $m\in\N$ we choose some $x_m\in \act^{-1}(\{y_m\})\cap D$. The resulting sequence $(x_m)$ is bounded since $D$ is bounded and thus has a convergent subsequence by the Bolzano-Weierstrass theorem. Without loss of generality we may assume 
$x_m\rightarrow x\in \overline{D}$ as $m\rightarrow\infty$.
Then by continuity, $\act(x_m)\rightarrow\act(x)=y$ as $m\rightarrow\infty$. If $x\in\partial D$, we directly have $y\in \act(\partial D)$. 
For the remaining case that $x\in D^\circ$, we use a homotopy type argument, wherein, as elsewhere in this work, functions of a scalar variable are applied to vectors coordinate-wise without using an extra notation. 
Let 
\[H(x,\lambda):=(1-\lambda) x +\lambda \act(x) \ \mathrm{for} \ \lambda \in  [0,1].\]
Then $H(x,0)=x$ and $H(x,1)=\act(x)$ and it can be directly verified that  $(x,\lambda) \mapsto H(x,\lambda)$ is continuous. The monotonicity of $\act$ implies that for every fixed $\lambda\in  [0,1]$, the mapping $x\mapsto H(x,\lambda)$ is coordinate-wise monotonically increasing and for $\lambda<1$ is coordinate-wise strictly monotonically increasing. We thus have that for fixed $\lambda\in [0,1)$, the mapping $x\mapsto H(x,\lambda)$ is a homeomorphism between the compact sets $\overline{D}$ and $H(\overline{D},\lambda)$. 
The continuity of $H$ together with the fact that the $\act(x_m)$ converge to $y$ imply that for every sequence $(\lambda_m)$ in $[0,1)$ with $\lambda_m\rightarrow 1$ as $m\rightarrow \infty $, we have $H(x_m,\lambda_m)\rightarrow y$ as $m\rightarrow \infty$.
Let us fix such a sequence $(\lambda_m)$ for the sequel of the proof.
Since $D^\circ$ is open and $x\in D^\circ$, the fact that $x\mapsto H(x,\lambda_m)$ is a homeomorphism on $\overline{D}$ for every $m$ implies that $H(D^\circ,\lambda_m)$ is open, so that $H(x_m,\lambda_m)$ is automatically an interior point of $H(D^\circ,\lambda_m)$. With a similar argument we have $\partial H(D,\lambda_m)=H(\partial D,\lambda_m)$ for all $m\in \N$, so that
\[\varepsilon_m:=\dist(H(x_m,\lambda_m), H(\partial D,\lambda_m))=\dist(H(x_m,\lambda_m),\partial H(D,\lambda_m)).\]
If there were a lower bound $\varepsilon_m\geq \delta>0$ for all $m\in\N$, we would have $U_\delta(H(x_m,\lambda_m))\subset H(D,\lambda_m)$ for all $m\in \N$, so that by continuity of $H$
\begin{equation*}
 U_\delta(y) = U_\delta(H(x,1))\subset H(D,1)=\act (D).
\end{equation*}
But this contradicts the fact that $y\in \partial \act(D)$. 
The previous argument can be applied to every subsequence and it thus follows that $\varepsilon_m\rightarrow 0$ as $m\rightarrow\infty$. Again by continuity of $H$, we obtain  
\[\dist(H(x_m,\lambda_m),H(\partial D,1))=\dist(H(x_m,\lambda_m),\act(\partial D))\rightarrow 0\ \mathrm{as}\ m\rightarrow \infty.\]
Since $\lim\limits_{m\rightarrow\infty}H(x_m,\lambda_m)=y$,
we have shown that $y\in \act(\partial D)$.

\end{proof}

\begin{proof} (Theorem \ref{attractor_unbounded})
To show the result by contradiction, we assume that for some attractive point $x^*$, the basin of attraction $B(x^*)$ has a bounded component $C$. By the definition of connectivity, we can find an open set $D$ such that $C\subset D$ and $D\cap (B(x^*)\setminus C)=\emptyset$, hence $\partial D\subset \R^{n_0}\setminus B(x^*)$, i.e. the points on $\partial D$ are not attracted by $x^*$. Since $C$ is bounded we can assume $D$ to be bounded. Let us define  $D_0:=D$, $D_k:=F_k(D)$, and, for some $x\in C$, $x_k =F_k(x)$ for $k=1,...,L$. Let us assume that $W_{k+1}$, where $0\leq k<L$, is the first weight matrix with rank strictly less than $n_0$. Such a weight matrix is assumed to exist in the statement of theorem, c.f. (\ref{rankCond}). 

We first consider the case that $k>0$.
According to our choice of $k$, we have that the weight matrices $W_1,...,W_k$ have full rank, and as their number of rows (the width of the network) is bounded by $n_0$, it iteratively follows (from $W_1$ to $
W_k$) that these are square matrices. The corresponding affine mappings $\A_j$, $j=1,...,k$, are hence homeomorphisms and thus $\A_j(\partial D_{j-1})=\partial\A_j(D_{j-1})$ for all $j\leq k$. 
With Lemma \ref{lemma_nottoouter} we can thus conclude that $\partial\act(\A_j(D_{j-1}))\subset \act(\A_j(\partial D_{j-1}))$ for $j=1,...,k$, and hence 
\begin{equation}\label{F_kPartial_01}
    \partial D_k=\partial F_k(D)\subset F_k(\partial D).
\end{equation}

Next, in the case that $k=0$, note that (\ref{F_kPartial_01}) holds if we set $F_0$ to be the identity map.

We now use the fact that $\rank(W_{k+1})<n_0$. Taking into account that $W_k\in\R^{n_0\times n_0}$, which implies that $W_{k+1}$ has $n_0$ columns, this implies that we can find a $v\neq 0$ in the null space of $W_{k+1}$. Considering further that $x_k\in D_k=F_k(D)$, (\ref{F_kPartial_01}) implies that we can choose a $\lambda \in \R$ such that 
\begin{equation}\label{intersectWithin}
y_k=x_k+\lambda v \in F_k(\partial D)
\end{equation}
as $F_k(\partial D)$ is compact, thus bounded. Then $W_{k+1} x_{k} +b_{k+1}=W_{k+1} y_{k} +b_{k+1} $ so that we can choose a $y\in F_k^{-1}(\{y_k\})\cap \partial D$ with $F(y)=F(x)$. This contradicts $\partial D\cap B(x^*)=\emptyset$.
\end{proof}

\begin{proof}(Theorem \ref{attractor_doesnot_enclosed_points})
To show the assertion by contradiction, let us assume that for some attractive point $x^*$, there exists a bounded component $C$ of $\R^{n_0}\setminus B(x^*)$. Then there is an open set $D\subset \R^{n_0}$ such that $C\subset D$ and $D\cap \R^{n_0}\setminus (B(x^*)\cup C)=\emptyset$. This gives $\partial D \subset B(x^*)$ where $\partial D$ can be assumed to be compact since $C$ is bounded. 

For the case that all weight matrices are full-rank, it follows from the fact that their number of rows (the width) is upper bounded by $n_0$ that they are square matrices. Thus, in this case the $\A_j$ for $j=1,...,m$ are homeomorphisms and hence map boundary points to boundary points. Together with Lemma \ref{lemma_nottoouter} this can be applied iteratively to all layers and finally gives $\partial F(D)\subset F(\partial D)$. Thus for the $k$-fold application of $F$ we have $\partial F^k(D)\subset F^k(\partial D)$ for all $k\in\N$. Taking into account that $\partial D\subset B(x^*)$, we can conclude that for sufficiently large $k$ and sufficiently small $\varepsilon>0$
\[\partial F^k(D)\subset F^k(\partial D)\subset U_\varepsilon(x^*)\subset B(x^*).\]
But according to $F(C)\subset F(D)$, this implies $F^k(C)\subset  U_\varepsilon(x^*)\subset B(x^*)$, which is impossible since $C$ is assumed to be a component of $\R^{n_0}\setminus B(x^*)$. 

In the remaining case we have that at least one weight matrix has rank strictly less than $n_0$. Let $W_j$ be the first matrix that obeys this condition. Then, as in the proof of Theorem \ref{attractor_unbounded}, the iterative application of Lemma \ref{lemma_nottoouter} to the first $j-1$ layers gives 
\begin{equation}\label{theo2inclusion}
   \partial F_{j-1}(D)\subset F_{j-1}(\partial D),
\end{equation}
where $F_0$ means the identity map for if $j=1$.
In the same way as argued to obtain (\ref{intersectWithin}) in the proof of Theorem \ref{attractor_unbounded}, we can conclude that there is a $v\neq 0$ in the null space of $W_j$, so that for some arbitrary $x\in C$, there is a $\lambda\in\R$ with
\[ F_{j-1}(x)+\lambda v=y\in \partial F_{j-1}(D).\]
This implies $W_jF_{j-1}(x)=W_jy$, so that by (\ref{theo2inclusion}) and $\partial D\subset B(x^*)$, we have that $F^k(x)\rightarrow x^*$ as $k\rightarrow \infty$. This contradicts $C\cap B(x^*)=\emptyset$.
\end{proof}

The following example shows that Theorem \ref{attractor_unbounded} does not hold without the condition (\ref{rankCond}).
\begin{example}\label{ex:RankCond}
Let 
\[
\act(x)=\begin{cases}
2x+1/4 \ \ &\mathrm{for }\, x<0\\
1/4 \ \ &\mathrm{for }\, x\in [0, 1]\\
2(x-1)+1/4  \ \ &\mathrm{for }\, x>1,
\end{cases}
\]
then $\act(x)$ has fixed points at $x=-1/4, x=1/4,\, x=7/4$ and 
\[
\lim\limits_{m\rightarrow \infty} \act^m(x)=\begin{cases}
-\infty \ \ &\mathrm{for }\, x<-1/4\\
1/4 \ \  &\mathrm{for }\, x\in (-1/4, 7/4)\\
+\infty  \ \ &\mathrm{for }\, x>7/4.\\
\end{cases}
\]
Hence, with $W_2=W_1=I_2$ the autoencoder $x\mapsto W_2\act(W_1x)$ has $(-1/4, 7/4)\times (-1/4, 7/4)$ as the basin of attraction of $x^*=(1/4,1/4)$.
\end{example}

In the next example, we construct an autoencoder from $\R$ to $\R$ of width two, that has a disconnected basin of attraction. As it maps from $\R$ to $\R$, the latter implies that the complement of the said basin of attraction has a bounded component. The example thus shows that the width condition in Theorem \ref{attractor_unbounded}, Theorem \ref{attractor_doesnot_enclosed_points} cannot be dropped.
\begin{example}\label{ex:unboundBasin}
Let \[
\act(x)=\begin{cases}
1/4x \ \ &\mathrm{for }\, x<0\\
x \ \  &\mathrm{for }\, x\geq 0
\end{cases}
\]
be a leaky ReLU activation function and let
\[f_1(x)=\act(2(x-1))+3/4 \ \mathrm{and}\ \ f_2(x)=-\act(3(x-5))-27/8.
\]
Then $f=f_1+f_2$ has fixed points at $x^*=1/2$, $y^*=7/2$ and $z^*=83/16$. The derivative of $f$ in a neighborhood of $x^*$ equals $-1/4$ which shows that $x^*$ is an attractive fixed point, c.f. \cite{strogatz2018nonlinear}. Since $y^*,\, z^*$ are themselves fixed points we have $y^*,\, z^*\notin B(x^*)$. However, for $\tilde{x}=79/8$ we have $f(\tilde{x})=1/2$ and hence $\tilde{x}\in B(x^*)$. Without figuring out the exact basin of attraction $B(x^*)$, we can conclude that $B(x^*)$ is not path-connected and, as $x^*<y^*< z^*<\tilde{x}$, the complement of $B(x^*)$ has a bounded component. It is directly seen that $f$ can be written as a network function of width two and depth two.
\end{example}


\subsection{Path-connectivity of basins of attraction}

In this subsection, we first give an example of an autoencoder $F:\R^2\rightarrow \R^2$ of depth six and width two, that has a disconnected basin of attraction. 
Conversely, we show in Theorem \ref{attractor_pathconnected} that path-connectivity of basins of attraction is guaranteed for neural networks with width equal to the input dimension when the activation function is monotonic and surjective and the weight matrices are all invertible. 

\begin{figure}[ht]
	\centering
	\includegraphics[scale=0.5]{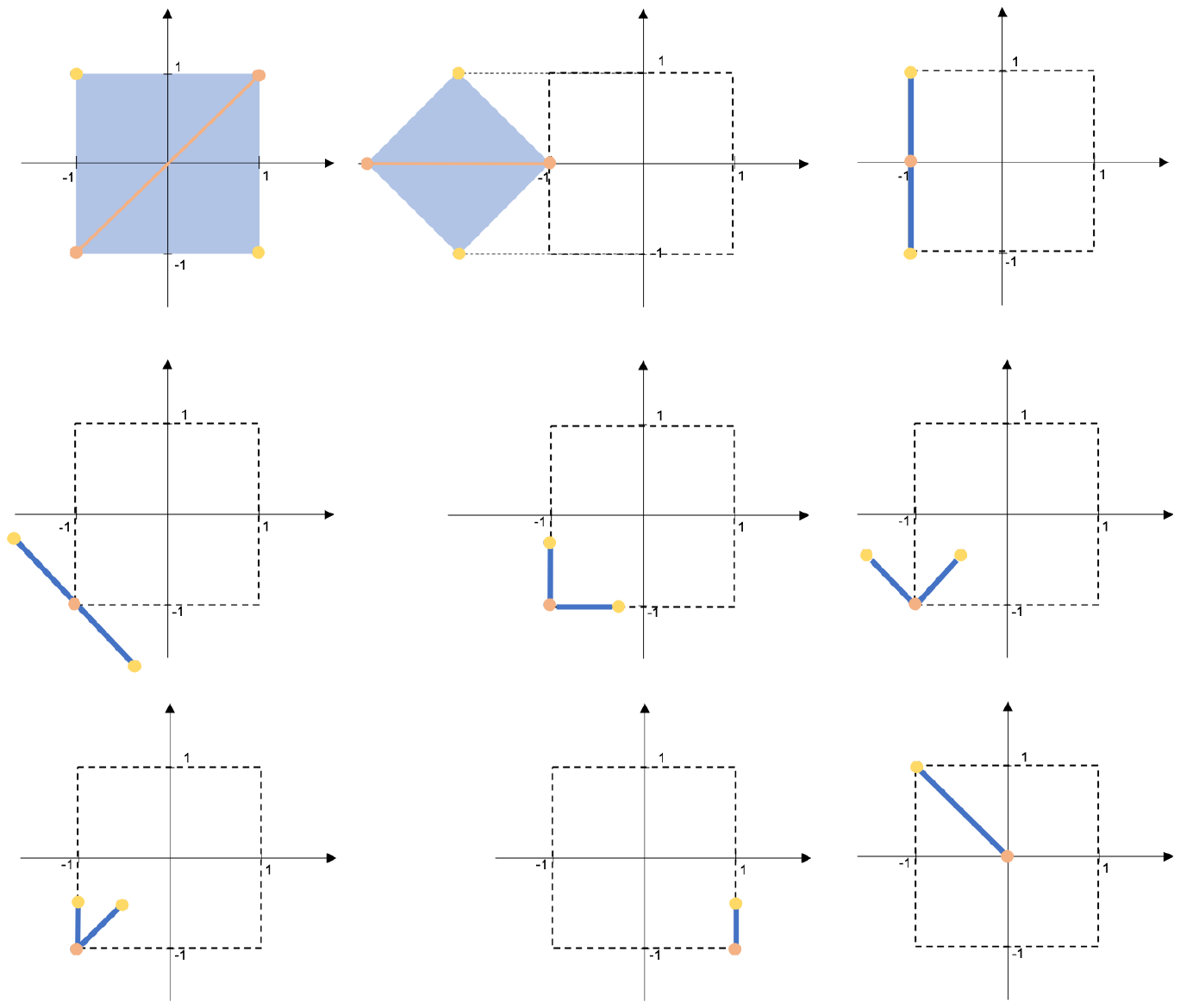} 
	\caption{Input progression through the layers of $F$ from Example \ref{UnconBasinEx}. Yellow sets/points depict the range of $S_1,S_3$, red sets/points depict the range of $S_2,S_4,S_{2,4}$, c.f. Eq. (\ref{setsS}), and blue depicts the range of the input space $\R^2$. The blue set in the different sub figures show the following: 
	Upper row, from left to right: $F_1(\R^2)$ Eq. (\ref{Q_1}), $\A_2(Q_1)$ Eq. (\ref{Q_2_1}), $F_2(\R^2)$ Eq. (\ref{Q_2});
	Middle row, from left to right: $\A_3(Q_2)$ Eq. (\ref{Q_3_1}), $F_3(\R^2)$ Eq. (\ref{Q_3}), $\A_4(Q_3)$ Eq.(\ref{Q_4_1});
	Lower row, from left to right: $F_4(\R^2)$ Eq. (\ref{Q_4}), $F_5(\R^2)$ Eq.(\ref{Q_5}), $F(\R^2)$}
	\label{fig:evolutionQ}
	
\end{figure}

\begin{example}\label{UnconBasinEx}
The target is the construction of a non path-connected basin of attraction for $x^*:=(-1,1)^T$. We refer to Figure \ref{fig:evolutionQ} for a depiction of the following construction. 
Let \[
\act(x)=\begin{cases}
-1 \ \ &\mathrm{for }\ x<-1\\
x \ \  &\mathrm{for }\ x\in [-1,1]\\
1 \ \ &\mathrm{for }\ x>1.
\end{cases}
\]
be the activation function, which is also known as \textit{HardTanh} function. 
Furthermore, we would like to remind the readers of the notation we introduced in (\ref{networkFun}) and  (\ref{networkFunLin}).
We define $W_1:=2I_2$ (recall that $I_2$ is the $2\times 2$ identity matrix) and $b_1:=(0,0)^T$. The application of $2I_2$ expands $[-1,1]\times [-1,1]$ to $[-2,2]\times [-2,2]$ which implies that $x\mapsto \act(W_1x)$ maps the points of an open neighborhood of $x^*$ to $x^*$, which is finally needed to have a basin of attraction in accordance with Definition \ref{defAttPoint}. We then have 
\begin{equation}\label{Q_1}
F_1(\R^2)=[-1,1]\times [-1,1]=:Q_1.
\end{equation}
And for 
\begin{equation}\label{setsS}
    \begin{array}{lll}
    S_1&:=&\{(x_1,x_2)^T\in\R^2: x_1\leq  -1/2 \ \mathrm{ and } \  x_2\geq1/2\}\\
    S_2&:=&\{(x_1,x_2)^T\in\R^2: x_1\geq 1/2 \ \mathrm{ and } \  x_2\geq 1/2\}\\
    S_3&:=&\{(x_1,x_2)^T\in\R^2: x_1\geq1/2 \ \mathrm{ and } \  x_2\leq -1/2\}\\
    S_4&:=&\{(x_1,x_2)^T\in\R^2: x_1\leq  -1/2 \ \mathrm{ and } \  x_2\leq -1/2\}\\
    S_{2,4}&:=&\conv((-1/2,-1/2)^T,(1/2,1/2)^T),
    \end{array}
\end{equation}
we have $F_1(S_1)=(-1,1)^T$, $F_1(S_2)=(1,1)^T$, $F_1(S_3)=(1,-1)^T$, $F_1(S_4)=(-1,-1)^T$ and $F_1(S_{2,4})=\conv((-1,-1)^T,(1,1)^T)$.
For the next steps of this example, let $R_\alpha\in\R^{2\times 2}$ be the rotation matrix,
that rotates a vector counterclockwise by an angle of $\alpha$ and preserves its length. 
Let $W_2:=2^{-1/2}R_{-\pi/4}$ and $b_2:=(-2,0)^T$, then
\begin{equation}\label{Q_2_1}
\A_2(Q_1)=\{(x_1,x_2)^T\in\R^2: \Vert x_1+2\Vert +\Vert x_2\Vert \leq 1\},
\end{equation}
so that
\begin{equation}\label{Q_2}
F_2(\R^2)=\conv((-1,1)^T,(-1,-1)^T)=:Q_2.
\end{equation}
Let us summarize the the status of the construction. The following hold true
\begin{equation*}
S_1 \subset  F_2^{-1}(\{(-1,1)^T\}),\ \ 
S_2\cup S_4\cup S_{2,4}\subset  F_2^{-1}(\{(-1,0)^T\}),\ \
S_3\subset  F_2^{-1}(\{(-1,-1)^T\}).
\end{equation*}
We also observe that the inverse images $F_2^{-1}(\{(-1,1)^T\})$ and $F_2^{-1}(\{(-1,-1)^T\})$ are not path-connected since they are separated by $S_2\cup S_4\cup S_{2,4}$.
Our next goal is to map the endpoints of $Q_2$ to a common point without compressing the whole line segment to that point.
To this end let $W_3:=R_{\pi/4}$ and $b_3:=R_{\pi/4}\,(1,0)^T+(-1,-1)^T$. Then
\begin{equation}\label{Q_3_1}
\A_3(Q_2)=\conv\left(\begin{pmatrix}-2^{-1/2}-1\\2^{-1/2}-1\end{pmatrix},\begin{pmatrix}2^{-1/2}-1\\-2^{-1/2}-1\end{pmatrix}\right),
\end{equation}
and from there one can conclude that $F_3(\R^2)=:Q_3=Q_{3,1}\cup Q_{3,2}$ with
\begin{equation}\label{Q_3}
Q_{3,1}:=\conv\left(\begin{pmatrix}-1\\2^{-1/2} -1\end{pmatrix},\begin{pmatrix}-1\\-1\end{pmatrix}\right), \ Q_{3,2}:=\conv\left(\begin{pmatrix}-1\\-1\end{pmatrix},\begin{pmatrix}2^{-1/2}-1\\-1\end{pmatrix}\right).
\end{equation}
Next, set $W_4:=R_{\pi/4}$ and $b_4:=R_{\pi/4}\,(1,1)^T+(-1,-1)^T$, which gives 
\begin{equation}\label{Q_4_1}
\A_4(Q_3)=\conv\left(\begin{pmatrix}-3/2\\-1/2\end{pmatrix},\begin{pmatrix}-1\\    -1\end{pmatrix}\right)\, \bigcup \, \conv\left(\begin{pmatrix}-1 \\ -1\end{pmatrix},\begin{pmatrix}-1/2 \\ -1/2\end{pmatrix}\right).
\end{equation}
And we then have $F_4(\R^2)=:Q_4=Q_{4,1}\cup Q_{4,2}$, where
\begin{equation}\label{Q_4}
Q_{4,1}:=\conv\left(\begin{pmatrix}-1\\ -1/2\end{pmatrix},\begin{pmatrix}-1\\ -1\end{pmatrix}\right),\\ 
Q_{4,2}:=\conv\left(\begin{pmatrix}-1\\ -1\end{pmatrix},\begin{pmatrix}-1/2\\-1/2\end{pmatrix}\right).
\end{equation}
For the fifth layer define $W_5:=I_2$ and $b_5:=(2,0)^T$, so that
\begin{equation}\label{Q_5}
F_5(Q_4)=\conv\left(\begin{pmatrix}1\\ -1\end{pmatrix},\, \begin{pmatrix}1\\ -1/2\end{pmatrix}\right)=:Q_5. 
\end{equation}
Let us summarize again the status of the construction:
\begin{equation*}
    S_1\cup S_3 \subset  F_5^{-1}(\{(1,-1/2)^T\}), \ S_2\cup S_4\cup S_{2,4}\subset  F_5^{-1}(\{(1,-1)^T\}).
\end{equation*}
The final layer shall be defined to map $Q_5$ to the line segment that connects $(-1,1)^T$ with the origin, where $(1,-1)^T$ is mapped to $(0,0)^T$ and  $(1,-1/2)^T$ is mapped to $(-1,1)^T$. This is accomplished by defining $W_6:=2^{3/2}R_{\pi/4}$, $b_6:=2^{3/2}R_{\pi/4}\,(-1,1)^T$. The final network function is then given by $F(x):=W_6F_5(x)+b_6$.

\end{example}
From the construction in the latter example it turns out that:
\begin{enumerate}
    \item $F(x^*)=x^*$ and $S_1\cup S_3\subset B(x^*)$.
    \item $S_2\cup S_4\cup S_{2,4}\subset  F^{-1}(\{(0,0)^T\})$ and since $(0,0)^T$ is also a fixed point of $F$, we can conclude $(S_2\cup S_4\cup S_{2,4})\,\cap B(x^*)=\emptyset$. 
    \item The latter shows that $B(x^*)$ cannot be path-connected since points from $S_1$ are separated from points in $S_3$ by $S_2\cup S_4\cup S_{2,4}$.
    \item The remaining set $S_0:=\R^2\setminus (S_1\cup S_2 \cup S_3\cup S_4 \cup S_{2,4})$ is mapped by $F$ to the  open line segment $\tilde{Q}:=\conv((-1,1)^T,(0,0)^T)\setminus \{(-1,1)^T,(0,0)^T\}$. A subsequent application of $F$ then maps the points in $\conv((-1,1)^T,(-1/2,1/2)^T)$ to $x^*$. The iterative application of this reasoning shows that every point in $\conv((-1,1)^T,(0,0)^T)$ is attracted by $x^*$, except for $(0,0)^T$ since $F((0,0)^T)=(0,0)^T$. This shows that $F^k(x)\rightarrow x^*$ as $k\rightarrow\infty$ for all $x\in S_0$ and thus $S_0\subset B(x^*)$. 
    \item It is clear that appending a layer to $F$, wherein the ($2\times 2$)-weight matrix has diagonal entries equal to $1/2$ and off diagonal entries equal to $-1/2$ (and hence has rank one) and with zero bias will give the same mapping as it does not affect $F(\R^2)$. This shows that in the situation of Theorem \ref{attractor_unbounded}, the basin of attraction can have several components.
\end{enumerate}

The following result shows that a construction as in Example \ref{UnconBasinEx} is only possible for non-surjective activation functions.

\begin{theorem}\label{attractor_pathconnected}
Let $F$ be an autoencoder with a surjective, monotonically increasing activation function $\act$, and let all $W_j$, $j=1,...,L$, be full rank, square matrices. If $x^*$ is an attractive point of $F$, then $B(x^*)$ is path-connected.
\end{theorem}

The conditions of the above result are motivated by the investigations in \cite{nguyen2018neural}, wherein, for pyramidal structured neural networks with strictly monotonic and surjective activation functions, the corresponding decision regions are shown to be connected. 
Note that our assumptions on the activation function are weaker than in \cite{nguyen2018neural}. 

The proof is prepared by three auxiliary results.

\begin{lemma}\label{lemma_sigma_x1x2inUy}
Let $x_1,x_2\in \R^n$ and $\act:\R\rightarrow \R$ monotonic such that 
$ \act(x_1)\in U_\varepsilon(y) \text{ and } \act(x_2)\in U_\varepsilon(y)$
for some $y\in\R^n$ and $\varepsilon>0$. Then $\act(x)\in U_{\varepsilon \sqrt{n}}(y)$ for every $x\in \conv(x_1,x_2)$.
\end{lemma}

\begin{proof}
By the monotonicity of $\act $, we have for every $\lambda\in [0,1]$ and every coordinate $j\in\{1,...,n\}$
\begin{equation}\label{uneq_mono}
\min\{(\act(x_1))_{(j)},(\act(x_2))_{(j)} \}\leq \left(\act(\lambda x_1+(1-\lambda )x_2)\right)_{(j)}\leq \max\{(\act(x_1))_{(j)},(\act(x_2))_{(j)} \}.
\end{equation}
Since by assumption, for all $j\in\{1,...,n\}$ we have $\Vert (\act(x_1))_{(j)}-y_{(j)}\Vert <\varepsilon$ and $\Vert (\act(x_2))_{(j)}-y_{(j)}\Vert <\varepsilon$, we can conclude from (\ref{uneq_mono}) that 
\begin{equation*}
  \Vert \left(\act(\lambda x_1+(1-\lambda )x_2)\right)_{(j)}-y_{(j)}\Vert < \varepsilon  
\end{equation*}
 for $j=1,...,n$ and hence, by the definition of the euclidean norm, we obtain $ \Vert \act(\lambda x_1+(1-\lambda )x_2)-y\Vert < \varepsilon \sqrt{n}$.
\end{proof}

\begin{lemma}\label{lemma_sigma_inversImagePathCon}
Let $\act:\R\rightarrow\R$ be surjective and monotonic, and $D\subset \R^n$ open and path-connected. 
Then the inverse image $\act^{-1}(D)$ (under the coordinate-wise application of $\act$) is path connected.
\end{lemma}
\begin{proof}
For $x_1,x_2\in \act^{-1}(D)$ let $y_1=\act(x_1),\, y_2=\act(x_2)$ and $\gamma:[0,1]\rightarrow D$ a continuous path with $\gamma(0)=y_1,\, \gamma(1)=y_2$. Let $\varepsilon>0$ be so small that $U_{\varepsilon \sqrt{n}} (\gamma(t))\subset D$ for all $t\in [0,1]$, which is possible since $\gamma([0,1])$ is compact and $D$ is open. Consider a partitioning $0=t_1<t_2<...<t_m=1$ such that $\Vert \gamma(t_j)-\gamma(t_{j+1})\Vert <\varepsilon$, for $j=1,...,m-1$. Since $\act(\R)=\R$ we have that $\act$ (applied coordinate-wise) is surjective from $\R^n$ to $\R^n$, which implies that for every $j\in \{2,...,m-1\}$ we can choose some $v_l\in \act^{-1}(\{\gamma(t_j)\})$. We further set $v_1:=x_1\in\act^{-1}(\{y_1\})$ and $v_m:=x_2\in\act^{-1}(\{y_2\})$. Then by construction, Lemma \ref{lemma_sigma_x1x2inUy} yields that the range of $\conv(v_j,v_{j+1})$ under $\act$ is a subset of $U_{\varepsilon \sqrt{n}}(\gamma(t_j))\subset D$, for $j=1,...,m-1$, showing that $\conv(v_j,v_{j-1})$ is contained in $\act^{-1}(D)$. The concatenation of the line segments $\conv(v_1,v_{2})$, $\conv(v_2,v_{3})$,..., $\conv(v_{m-1},v_{m})$ gives a continuous path connecting $x_1,x_2$ in $\act^{-1}(D)$.
\end{proof}

The following result is a direct consequence of linearity, but for the sake of clarity we fix the assertion in a lemma. 
\begin{lemma}\label{lemma_linear_inversImagePathCon}
Let $\A:\R^n\rightarrow\R^m$, $\A(x)=Wx+b$, where $W\in\R^{m\times n}$, $b\in\R^{m}$, be a surjective, affine mapping, and $D\subset \R^m$ open and path-connected. Then the inverse image $\A^{-1}(D)$ is path-connected.  
\end{lemma}
As this result is obvious, the proof is omitted.
\begin{proof}(Theorem \ref{attractor_pathconnected})
Let $x^*$ be an attractive point of $F$. Since $x^*$ is an interior point of $B(x^*)$, we can find an $\varepsilon>0$ such that $U_\varepsilon(x^*)\subset  B(x^*)$. Taking into account that all weight matrices are assumed to be square and have full rank, we deduce that we can apply Lemma \ref{lemma_linear_inversImagePathCon} to every $\A_j$, $j=1,...,L$. The iterative application of Lemma \ref{lemma_linear_inversImagePathCon} and Lemma \ref{lemma_sigma_inversImagePathCon} from the last layer to the first layer yields that $F^{-1}(U_\varepsilon(x^*))$ is path connected, and hence so are $F^{-k}(U_\varepsilon(x^*))\subset B(x^*)$ for all $k\in\N$. Thus, given $x_1, x_2\in B(x^*)$ and a common $k\in\N$ such that $F^k(x_1)\in U_\varepsilon(x^*)$ and $F^k(x_2)\in U_\varepsilon(x^*)$, we obtain that $x_1,x_2$ can be connected by a path in $F^{-k}(U_\varepsilon(x^*))\subset B(x^*)$.
\end{proof}

\subsection{Discussion}
Our above investigations in Section \ref{topPropertiesSec} are formally restricted to the case that $F^k(x)\rightarrow x^*$ as $k\rightarrow\infty$. If $(m_k)_{k\in\N}$ is some subsequence of the positive integers with $F^{m_k}(x)\rightarrow x^*$ as $k\rightarrow\infty$, then the question immediately arises whether the results from this Section \ref{topPropertiesSec} still hold true, when the definition of basin of attraction is generalised accordingly to 
\begin{equation}\label{basinGeneral}
    B(x^*,(m_k)_{k\in\N}):=\{x\in\R^{n_0}:\lim\limits_{k\rightarrow \infty}F^{m_k}(x)=x^*\}.
\end{equation}
The proofs in Section \ref{topPropertiesSec} can be carried out without difficulty in the similar way for such a case and we hence remark:
\begin{remark}\label{rem_subs}

For a fixed subsequence of positive integers $(m_k)$, and corresponding basins of attraction defined in Eq. (\ref{basinGeneral}), Theorem \ref{attractor_unbounded}, Theorem \ref{attractor_doesnot_enclosed_points} and Theorem \ref{attractor_pathconnected} still hold true.
\end{remark}
As the findings from \cite{radhakrishnan2020overparameterized} constitute the starting point of our investigation, we next place the results from Section \ref{topPropertiesSec} in the context of this work. 
\begin{enumerate}
    \item It is conjectured in \cite{radhakrishnan2020overparameterized} in their section entitled ``Discussion" that the partitioning produced by basins of attraction is closely related to the tessellation produced by 1-NN (nearest neighbor) classifiers.
    However, Theorem \ref{attractor_unbounded} states that, under the hypotheses of this result, basins of attraction cannot be bounded, whereas 1-NN neighborhoods can be bounded.
    Theorem \ref{attractor_doesnot_enclosed_points} and Theorem \ref{attractor_pathconnected} on the other hand show that, under the conditions of the respective results, basins of attraction and 1-NN neighborhoods (with respect to usual distances) are guaranteed to share the property of being path-connected and do not enclose a component of their respective complements.
    
    \item The assumptions of Theorem \ref{attractor_unbounded} and Theorem \ref{attractor_doesnot_enclosed_points} are satisfied in several experiments with fully-connected networks reported in \cite{radhakrishnan2020overparameterized}, c.f. Table 5, and (in appendix) Table S1, S2 therein. The conditions of Theorem \ref{attractor_pathconnected} are more restrictive and are not exactly met by the experiments reported in \cite{radhakrishnan2020overparameterized}. 
    
    \item Following \cite[Definition 2]{radhakrishnan2020overparameterized}, a \textbf{stable discrete limit cycle} for $f:\R^n\rightarrow\R^n$ is a set $X^*=\{x_1^*,...,x_m^*\}\subset\R^n$ with $f(x_j^*)= x^*_{(j\,\mathrm{mod}\, m)+1}$, $j=1,...,m$, and such that there exists an open neighborhood $\mathcal{O}$ of $X^*$, so that for every $x\in \mathcal{O}$, $f^{km}(x)$ converges to some point in $X^*$ as $k\rightarrow \infty$. In \cite{radhakrishnan2020overparameterized}, it is also empirically shown that autoencoders can be trained to produce stable discrete limit cycles $ X^*=\{x_1^*,...,x_m^*\}$ consisting of training data. Remark \ref{rem_subs} implies that Theorem \ref{attractor_unbounded}, Theorem \ref{attractor_doesnot_enclosed_points} and Theorem \ref{attractor_doesnot_enclosed_points} apply to corresponding basins of attraction $B(x_j^*,(mk)_{k\in\N})$ for $j=1,...,m$, c.f. (\ref{basinGeneral}).
\end{enumerate}
In contrast to conventional update rules in Hopfield networks, c.f. \cite{hopfield1982neural}, the following continuous update formula is proposed in \cite{ramsauer2020hopfield}: $\xi_{k+1}=X\, \mathrm{softmax}(\beta X^T  \xi_k)$. Therein, the attractive fixed points, referred to as the stored key pattern, are arranged column-wise in $X\in \R^{n_0\times N}$,  $\xi_0$ is the vector of input features and $\beta$ is a positive constant. The corresponding (Hopfield) network function $\xi\mapsto X\, \mathrm{softmax}(\beta X^T \xi)$ is similar to the network functions considered in this work. However, since softmax does not apply coordinate-wise, our results do not cover the case of those continuous Hopfield networks. 

\section{Approximation in the bounded width setting}\label{secDecReg}

In \cite{johnson2018deep}, it is shown that network functions $F:\R^{n_0}\rightarrow\R$ of width at most $n_0$ are not dense in the space of continuous functions on compact sets with non-empty interior with respect to uniform convergence, see also \cite{hanin2017approximating} and \cite{beise2018decision} for related results. Conversely, neural network functions of width larger than $n_0$ have universal approximation properties \cite{hanin2017universal,park2020minimum}.
In this section, we show that the arguments in the proofs in Section \ref{topPropertiesSec} allow us to derive a result that implies both \cite[Theorem 1]{johnson2018deep} and the lower bound of \cite[Theorem 1]{hanin2017approximating}.


\begin{theorem}\label{maxPrinc}
	Let $F:\R^{n_0}\rightarrow \R$ be a neural network of width not exceeding $n_0$ endowed with a continuous, monotonically increasing activation function $\act$, and let $D$ be some bounded subset of $\R^{n_0}$. Then
	\begin{equation*}
	    \max\{F(x):x\in \overline{D}\}=\max\{F(x):x\in \partial D\}.
	\end{equation*}
\end{theorem}
\begin{proof}
The arguments given in the proof of Theorem \ref{attractor_doesnot_enclosed_points} yield $\partial(F_{L-1}(D))\subset F_{L-1}(\partial D)$ for both cases, either $W_1,...,W_{L-1}$ are square and have full rank or that at least one of them has rank strictly less than $n_0$.
The assertion then follows since, as an affine function, $\A_L$ takes its maximum value on $ \partial(F_{L-1}(D))$.

\end{proof}
Note that, considering $-F$ instead of $F$, the latter result also implies that the minimum value is taken on $\partial D$.


Theorem \ref{maxPrinc} immediately gives the following.
\begin{corollary}
If $F$ and $D$ are as in Theorem \ref{maxPrinc}, and $F(x)=a$ for all $x\in \partial D$ and some fixed $a\in \R$, then $F(x)=a$ for all $x\in D$.
\end{corollary}

\section{Conclusion}
In this work, we derive topological properties of basins of attractions of autoencoders that have a width that does not exceed the input dimension. It is shown that in this setting, a basin of attraction is unbounded and its complementary set cannot have bounded components. The first of the latter results also requires that at least one weight matrix has rank strictly less than the input dimension. We also show that under certain stricter conditions, basins of attractions are always path-connected. By means of examples, it is further demonstrated that our conditions are necessary, or cannot be dropped without substitute. 
We argue that our results provide some first answers to a question formulated in \cite{radhakrishnan2020overparameterized} for the case of the neural network architectures taken under consideration herein.


\clearpage

\bibliographystyle{plain} 
\bibliography{literature.bib}

\end{document}